\pdfoutput=1

\documentclass[11pt]{article}

\usepackage{acl}

\usepackage{times}
\usepackage{latexsym}

\usepackage[T1]{fontenc}

\usepackage[utf8]{inputenc}

\usepackage{microtype}

\usepackage{inconsolata}

\usepackage{graphicx}

\usepackage{algorithm}
\usepackage{algorithmic}
\usepackage{multirow}
\usepackage{amsmath}
\usepackage{amssymb}
\usepackage{colortbl}
\usepackage{fontawesome5}
\definecolor{myyellow}{rgb}{1,1, 0.6}
\definecolor{myorange}{rgb}{1, 0.8, 0.6}
\definecolor{myred}{rgb}{1, 0.6, 0.6}
\definecolor{lightblue}{rgb}{0.8,0.9,1}
\usepackage{todonotes}

\usepackage{bm}
\renewcommand{\todo}[1]{\iffalse #1 \fi{\color{blue} \textbf{[TODO]}}}

\usepackage{makecell}
\usepackage{newfloat}
\usepackage{listings}

\title{\includegraphics[height=0.25in, width=0.25in]{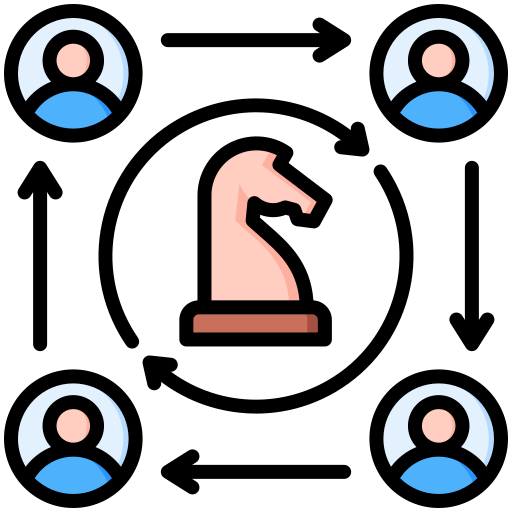} Align$^2$LLaVA: Cascaded Human and Large Language Model Preference Alignment for Multi-modal Instruction Curation}

\author{
    \textbf{Hongzhe Huang\textsuperscript{1}},
    \textbf{Jiang Liu\textsuperscript{1}},
    \textbf{Zhewen Yu\textsuperscript{1}},
    \textbf{Li Cai\textsuperscript{1}},
    \\
    \textbf{Dian Jiao\textsuperscript{1}},
    \textbf{Wenqiao Zhang\textsuperscript{1}$^{\dag}$},
    \textbf{Siliang Tang\textsuperscript{1}},
    \textbf{Juncheng Li\textsuperscript{1}},
    \\
    \textbf{Hao Jiang\textsuperscript{2}},
    \textbf{Haoyuan Li\textsuperscript{2}},
    \textbf{Yueting Zhuang\textsuperscript{1}}
    \\
    \\
    \textsuperscript{1}Zhejiang University,
    \textsuperscript{2}Alibaba,
    $^{\dag}$Corresponding Authors
}

\newcommand*{\method}{Align$^2$LLaVA}
\newcommand*{\dataset}{Align$^2$LLaVA-Instruct}

\begin{document}
\maketitle
\begin{figure*}
  \includegraphics[width=\textwidth]{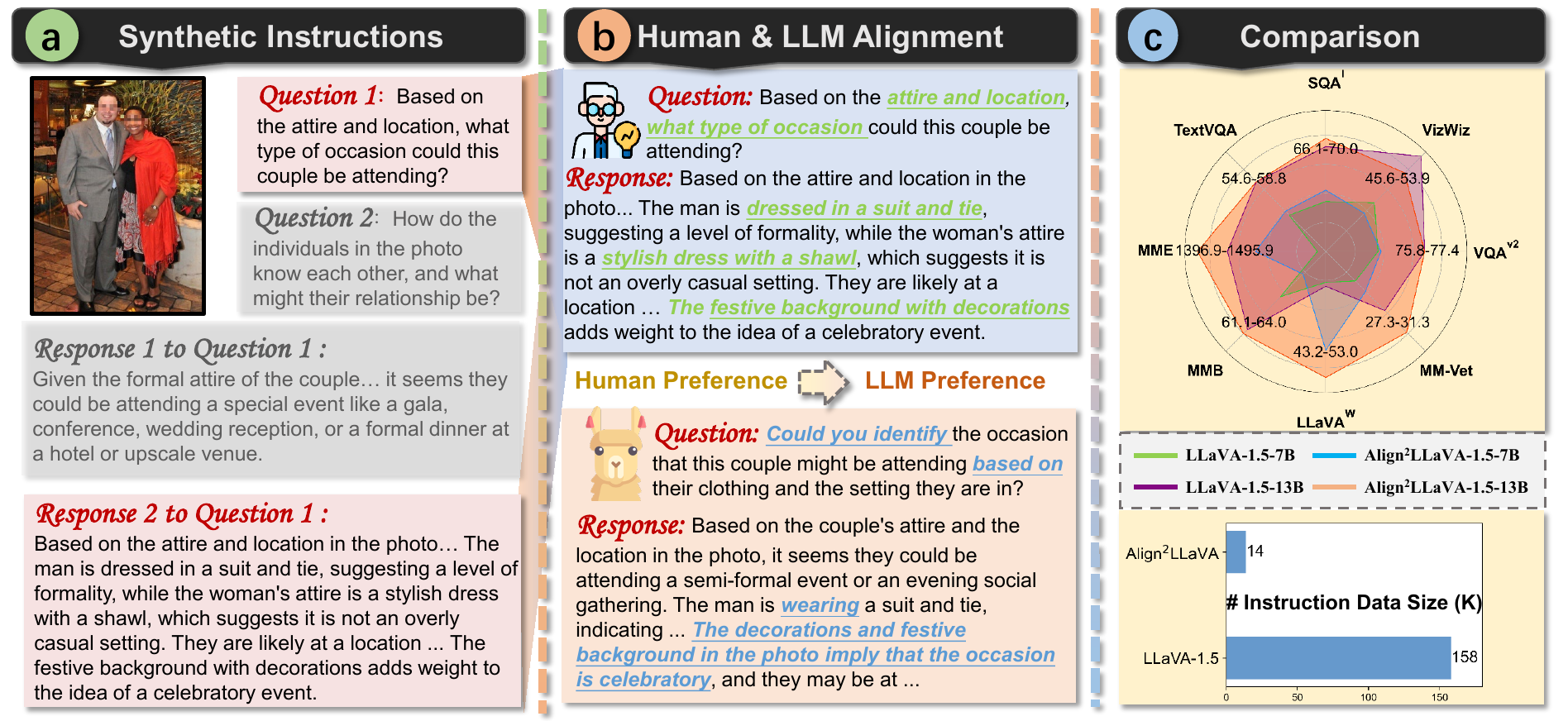}
  \caption{(a) An example of synthetic visual instruction data generated by LLMs. (b) Demonstration of  proposed cascaded human and LLM preference alignment. 
  (c) Across 8 benchmarks, our approach achieves comparable or superior performance to LLaVA-1.5 trained on the full dataset (top), using a significantly reduced instruction set (bottom), demonstrating the efficiency of our method.
  }
  \label{fig:intro}
\end{figure*}

\begin{abstract}
Recent advances in Multi-modal Large Language Models (MLLMs), such as LLaVA-series models, are driven by massive machine-generated instruction-following data tuning. Such automatic instruction collection pipelines, however, inadvertently introduce significant variability in data quality.
This paper introduces a novel instruction curation algorithm, derived from two unique perspectives,  \textbf{human and LLM preference alignment}, to compress this vast corpus of machine-generated multimodal instructions to a compact and high-quality form: (i) \emph{For human preference alignment}, we have collected a machine-generated multimodal instruction dataset and established a comprehensive set of both subjective and objective criteria to guide the data quality assessment critically from human experts. By doing so, a reward model was trained on the annotated dataset to internalize the nuanced human understanding of instruction alignment.
(ii) \emph{For LLM preference alignment}, given the instruction selected by the reward model, we propose leveraging the inner LLM used in MLLM to align the writing style of visual instructions with that of the inner LLM itself, resulting in LLM-aligned instruction improvement.
Extensive experiments demonstrate that we can \textbf{maintain or even improve model performance by compressing synthetic multimodal instructions by up to 90\%}. Impressively, by aggressively reducing the training instructions from 158k to 14k (9× smaller), our model consistently outperforms its full-size dataset counterpart across various MLLM benchmarks.  Our project is available at https://github.com/DCDmllm/Align2LLaVA.
\end{abstract}
\section{Introduction}
Recent triumphs in artificial intelligence have been driven notably by the development of large language models (LLMs)~\cite{ref:GPT4, ref:Gemini, ref:llama3}.  
Inspired by the impressive instruction-following capability of LLMs, visual instruction tuning~\cite{ref:LLaVA} has been proposed to extend LLMs into Multimodal LLMs (MLLMs) to perceive and
understand visual signals (\emph{e.g.}, images). 
Due to the scarcity of visual-language instruction-following data, contemporary MLLMs prefer to utilize the data reformation approach~\cite{ref:LLaVA}, which leverages text-only LLMs conditioned on image captions and bounding boxes to create synthetic multi-modal instructions involving visual content. 
Following the evolution, other works such as ALLaVA~\cite{chen2024allava} and VIGC~\cite{wangVIGCVisualInstruction2024}, have extended this automatic paradigm by utilizing visually-capable LLMs, demonstrating the potential value of utilizing synthetic instruction generation.

Despite their efforts and achievements, the simply synthetic instruction generation paradigm may result in sub-optimal multimodal instruction following capability, which can be summarized as
two key aspects: \textbf{(i) Inherently Noisy Instructions}:  The generated questions and corresponding responses of instructions may be incongruent or incomplete with the visual content, as they are produced by text-only large language models. 
Regarding question generation, common issues include irrelevance to the image, independence from visual cues, and repetitive queries in multi-turn dialogues. On the other hand, the corresponding responses may suffer from textual hallucinations or ignored visual details, as they are produced by text-only large language models. 
Figure \ref{fig:intro}(a) shows an example of the synthetic instructions. Of the two LLM-generated questions, the previous one demonstrates superior quality by incorporating queries relevant to image content, and providing subtle prompts to stimulate reasoning without revealing specific visual information. Conversely, the subsequent question lacks pertinence to the image, as it cannot be accurately answered based on the visual data provided. Moreover, comparing the two responses to the first question, the first one is notably deficient in visual details, as highlighted in the upper part of Figure \ref{fig:intro}(b). Such details are crucial for comprehensive and persuasive analysis.
These inappropriate questions and responses inevitably compromise the accurately visual-language alignments of the MLLMs.
\textbf{(ii) Internally Linguistic Gap}:  We refer the linguistic characteristic to the specific habits of vocabulary selection, grammar usage and sentence structuring used to express particular semantics. We argue that well-trained LLMs have unique writing style preferences, which are expressed in the output probabilities of candidate tokens when generating new tokens. 
In Figure \ref{fig:intro}(b), we present an instance of sentence writing manner gap. A comparison between the original instruction (top) and the LLM-revised instruction (bottom) reveals subtle stylistic variations, including word choice, sentence structure, and the degree of complexity in expressing the same idea. 
During the visual instruction tuning phase, the writing manner gap forces the LLM to change its original writing style, leading to performance degradation or even catastrophic forgetting.
Based on the aforementioned insights, a meaningful optimization goal of synthetic instruction generation is to effectively explore the well-aligned vision-instructions and minimize the linguistic gap, thereby boosting the multi-modal instruction-following capabilities.

Generally speaking, a prevalent strategy for quality enhancement of textually training data is \emph{data filtering}, \emph{e.g.}, employing pre-trained models to assess answer quality relative to given questions, subsequently retaining only instructions with high-quality answers~\cite{cao2023instruction, chen2023maybe05dataneeded, li2023quantity}. 
However, this paradigm is not the true savior for multi-modal instructions, mainly due to the current filtering methods being designed for text-only instruction data, \emph{i.e.}, lacking the reflect in depth of vision-language consistency among the image, question, and answer. Notably, these methods are trained on unfiltered data with noise and bias, which may lead to inaccurate assessments. What's more, current data filtering methods have not considered the linguistic gap between LLMs, handling the potential of synthetic instructions' potential.
Summing up, the premise of better multi-modal instruction filtering is to mitigate the aforementioned issues, pushing us to revisit the filtering solution for synthetic instructions.

To this end, we propose a novel data curation paradigm for synthetic multi-modal instructions, called \includegraphics[height=0.15in, width=0.15in]{pics/strategic.png} \textbf{\method{}}. Our bootstrapping philosophy is to progressively align human expert's and pre-trained LLM's preference, thereby respectively addressing the inherently noisy instructions and internally linguistic gap: \textbf{Step 1}, \emph{Human Knowledge Alignment}, this step is built upon the concept of integrating human feedback signals to comprehensively evaluate and filter both question and answer components of visual instructions. In detail, we collect human preference instructions emphasizing distinct quality criteria at question- and response-aligned level given images, \emph{i.e.}, respectively image-$k$ questions and question-$l$ responses datasets.
Human annotators independently rank each part based on metrics including image relevance, question comprehensiveness, answer accuracy, etc. 
By doing so, human-aligned instructions are obtained and thereafter employed to train two separate reward models for evaluation. Finally, utilizing the trained reward models, a two-stage filtering process is then applied to large-scale synthetic instructions, preserving examples that optimally conform to human-established quality standards within both question and answer domains. \textbf{Step 2}, \emph{LLM Characteristic Alignment}, we leverage the inner LLM to align the writing manner of soft-format visual instructions with that of the inner LLM itself, ensuring the original semantics of these instructions remain unchanged. The answer part of soft-format visual instructions is first rewritten by the inner LLM to match its writing manner, and then reviewed by the inner LLM to ensure the alignment of writing manner while preserving the original meaning. If the revised answer
is deemed unqualified during the review, the original answer is retained. 
By combining this two-step aligning process, the proposed synthetic multi-modal instruction dataset is created.

Our experiments empirically validated the efficacy of the proposed data curation method. We applied the pipeline to a 158K synthetic instruction dataset, resulting in a compressed dataset, \textbf{\dataset{}}. \dataset{} consists of $30\%$ of the original questions and $30\%$ of the remaining answers, which is $9\%$ of the full size. We subsequently fine-tuned the LLaVA-1.5~\cite{ref:llava1_5} model using \dataset{} combined with the caption and short VQA parts of the LLaVA-1.5-665K~\cite{ref:llava1_5} dataset, achieving comparable or even superior performance to the same architecture trained on the full instruction data, across a suite of 8 benchmarks. These benchmarks encompassed a diverse range of tasks, spanning traditional academic challenges and comprehensive instruction-following scenarios.

In summary, our contributions are threefold: 
\begin{itemize}
    \item We introduce a novel pipeline, \textbf{\method{}}, to curate high-quality synthetic visual instruction data by sequentially aligning with human preferences and LLM characteristics.
    \item Utilizing the curation pipeline, we present an aligned synthetic visual instruction dataset, \textbf{\dataset{}}, distinguished by its high quality and compact size.
    \item Through extensive experiments, we demonstrate the effectiveness of the proposed curation pipeline and dataset for MLLM training.
\end{itemize}

\section{Related Work}

\subsection{Learning from Human Knowledge}
The integration of human knowledge is becoming increasingly pivotal in aligning model behavior with human intent~\cite{lee2023aligning, nakano2021webgpt, wu2023better, xu2024imagereward, zhang2024hive, yu2024rlhf, zhang2023compress}. A crucial aspect is training a reward model~\cite{macglashan2017interactive} to directly capture human preferences for the outputs generated by the model. Recently, researchers have proposed to use reinforcement learning~\cite{schulman2017proximal} to fine-tune language models~\cite{instructgpt} and diffusion models~\cite{lee2023aligning, xu2024imagereward, zhang2024hive} utilizing the signals from the reward model. Additionally, there are works that leverage human knowledge to directly enhance the quality of datasets~\cite{yu2024rlhf, zhang2023compress}. These works integrate human knowledge with image-text data to mitigate visual and textual biases present in large-scale image-text datasets.

\subsection{Instruction Data Selection}
Recent researches show that instruction tuning could enable large language models (LLMs) to be tailored to specific domains, tasks, or applications by providing explicit instructions or guidelines~\cite{wei2021finetuned}. In order to enhance the instruction-following abilities of LLMs, previous works mainly focus on increasing the data sizes through various strategies~\cite{honovich2022unnatural, wang2022self, taori2023stanford, kopf2024openassistant}. However, LIMA~\cite{zhou2024lima} illustrates that even a small number of constructed high-quality instructions could empower the model with a powerful instruction-following capability.

Inspired by LIMA~\cite{zhou2024lima}, Instruction Mining~\cite{cao2023instruction} adopts a linear quality rule and bag of indicators to evaluate the quality of instruction-following data. AlpaGasus~\cite{chen2023alpagasus} directly leverages an external LLM (ChatGPT) to score each instruction and then selects 9K Alpaca data with a threshold. While their model surpasses the performance of the official Alpaca model which is trained on the complete dataset, this approach may neglect the intrinsic abilities of the base model, relying excessively on external models. Li et al.~\shortcite{li2023quantity} present a self-guided approach for LLMs to independently identify and choose relevant instruction pairs from extensive open-source datasets and in their approach they introduce an Instruction-Following Difficulty (IFD) metric as a tool to identify gaps in a model’s responses versus its autonomous generation capability. 
DataTailor~\cite{datatailor} focuses on instruction selection through three criteria: informativeness, uniqueness, and representativeness.
\section{Method}

\begin{figure*}[t]
    \centering
    \includegraphics[width=1\linewidth]{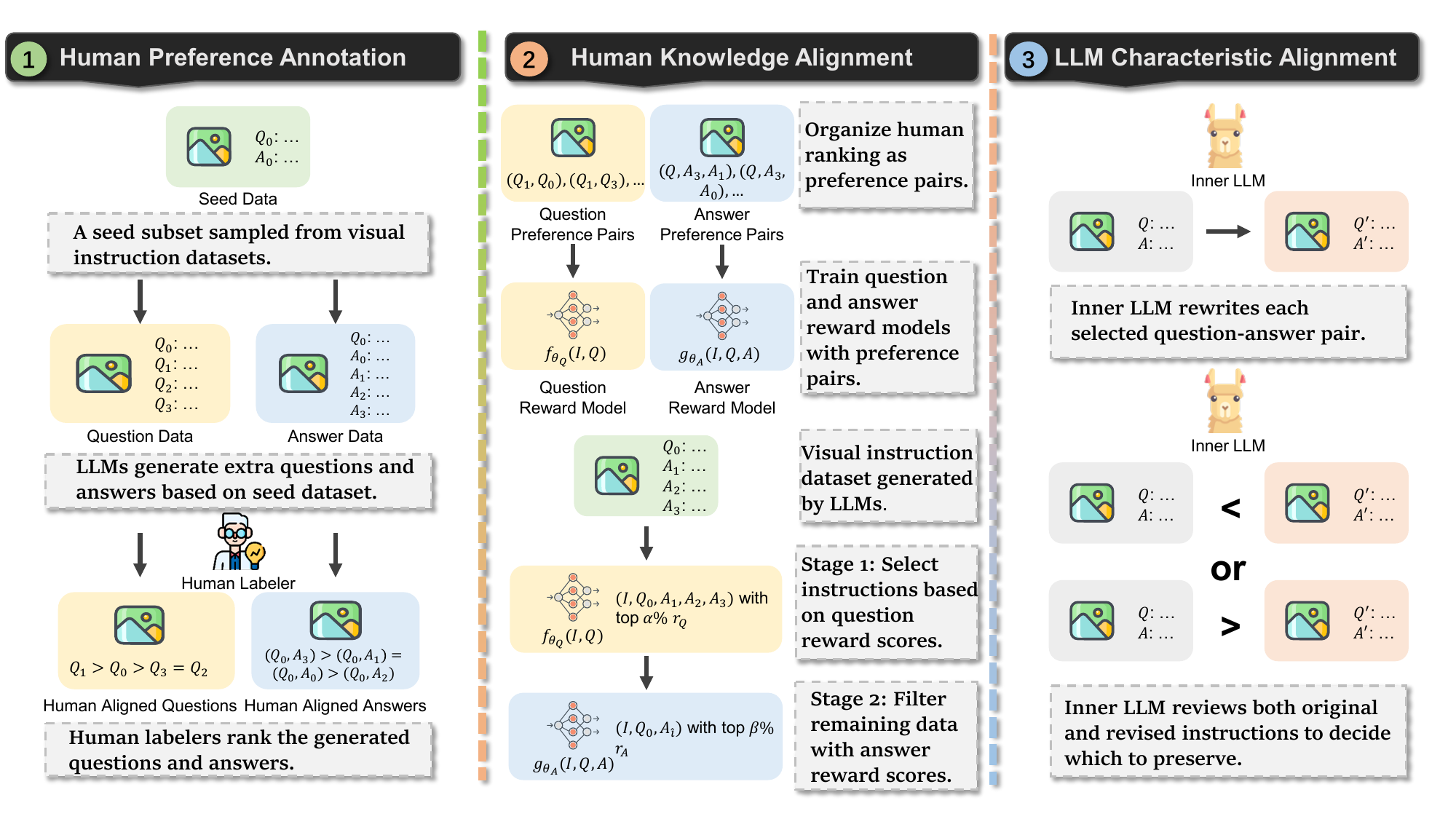}
    \caption{An overview of our data curation pipeline incorporating human knowledge and LLM characteristics. The process comprises three sequential steps: (1) Human preference data is curated through manual annotation on LLM generated questions and answers. (2) Two reward models are trained to align with human values, and subsequently utilized for large-scale data filtration. (3) An inner LLM is employed to rewrite and review the selected instructions.}
    \label{fig:method}
\end{figure*}

Our visual instruction curation pipeline comprises three sequential steps, as depicted in Figure \ref{fig:method}. Initially, two distinct datasets, enriched with human preferences, are constructed focusing on the question and answer components of instructions, respectively (Section \ref{method:pref_collect}). Leveraging these human-ranked datasets, two separate reward models are trained to align with human values, and subsequently serve as human-like evaluators to filter large-scale synthetic visual instruction data through a two-stage process (Section \ref{method:human_align}). 
Finally, an inner LLM is employed to refine the selected instructions in a rewrite-review manner, ensuring alignment with the LLM's writing style while preserving the original semantic content (Section \ref{method:LLM_align}).

\subsection{Human Preferences Collection} \label{method:pref_collect}

The initial step of our pipeline involves the curation of two distinct visual instruction datasets, each accompanied by corresponding human preference rankings. As depicted in Step 1 of Figure \ref{fig:method}, these datasets prioritize either the question or answer component of the instructions. The data collection process unfolds in two sequential phases: \emph{visual instruction generation} and \emph{human preference annotation}. The anticipated outcome is a question dataset ranked by comprehensiveness and visual relevance, as well as an answer dataset evaluated based on accuracy and answer alignment with the provided image and question.

\subsubsection{Visual Instruction Generation}

To initiate our pipeline, a seed dataset of 960 images is randomly selected from the LLaVA-instruct-158K~\cite{ref:LLaVA} dataset. Each image is paired with exactly one group of instructions generated by GPT-4~\cite{ref:GPT4}. Following LLaVA, these instructions are classified into three categories: \emph{detail description}, \emph{complex reasoning}, and \emph{conversation}, where only the \emph{conversation} samples incorporate multiple question-answer pairs. We also ignore the question part of \emph{detail description} instances, because they are randomly sampled from a fixed, limited set of pre-defined options.

Leveraging the seed dataset, we construct question and answer datasets with diverse instruction variations. To generate the question dataset, LLMs are prompted with images and tasked with producing multiple related queries, and images within the \textit{detail description} part are excluded from this dataset. For the answer dataset, LLMs are provided with both the original image and seed question to produce alternative responses. To enhance data diversity, both GPT-4V and text-only GPT-4 (following LLaVA's approach of incorporating MSCOCO~\cite{ref:mscoco} bounding boxes and captions) are employed. Each model generates three responses, with low-quality outputs (\emph{e.g.}, refusals with initial parts like "I'm sorry...") discarded. The seed questions or answers are also included in the datasets.

We finally collected a question dataset of 800 samples and an answer dataset of 960 samples. Formally, each seed sample is represented as $(I, Q_0, A_0)$, where $I$ is the image, $Q_0$ is the seed question, and $A_0$ is the seed answer. Question dataset samples are denoted as $(I, Q_0, Q_1, ..., Q_k)$, and and answer dataset samples as $(I, Q_0, A_0, A_1, ..., A_l)$, with $k, l \in [3, 6]$ representing the number of generated questions or answers, respectively.

\subsubsection{Human Preference Annotation}

To evaluate instruction quality, we employ 7 CS students as labelers. They are tasked with ranking each of the question and answer datasets according to distinct criteria, involving correctness, diversity, image-question-answer relevance, etc. 
Detailed annotation guidelines are provided in the appendix. In general, these criteria establish a multi-faceted evaluation framework that can assess visual instructions holistically. Moreover, to guarantee annotation accuracy and consistency, we have employed a multi-annotator approach. Each data sample is independently annotated by one annotator and subsequently reviewed by two or three additional annotators. 

\subsection{Human Knowledge Alignment} \label{method:human_align}

To align large-scale visual instruction data with human knowledge, we employ a human-preference-based filtration process utilizing curated datasets enriched with human feedback. Initially, two distinct reward models are trained separately on question and answer datasets to learn human preferences for these respective components. Subsequently, these trained reward models function as human-like evaluators to filter large-scale synthetic dataset across two sequential stages.

\subsubsection{Reward Model Training}

To provide a multi-dimensional assessment on visual instructions, we propose to train two distinct reward models on the human-ranked datasets, dedicated to evaluating the question and answer parts, respectively. By aligning these models with human preferences, we aim to cultivate a in-depth understanding of complex semantics and subtle nuances within instructions.

Following InstructGPT~\cite{instructgpt}, we structure human preferences as pairwise comparisons, and adopt a pairwise ranking loss for model training. Specifically, within the question dataset, each pair of questions associated with the same image are compared, resulting in a preference pair indicated by $(I, Q_i, Q_j)$ if $Q_i$ is superior to $Q_j$. Similarly, the answer dataset is organized as preference pairs $(I, Q_0, A_i, A_j)$, where $A_i$ is preferred over $A_j$ within the context of image $I$ and seed question $Q_0$. This yields at most $C_k^2$ question preference pairs and $C_l^2$ answer preference pairs for each image.

Subsequently, two reward models are independently trained on question and answer preference pairs, with loss functions defined as:
\begin{align}
loss_Q(\theta_Q) = &-\mathbb{E}_{(I, Q_i, Q_j) \sim \mathcal{D}^\mathrm{HF}_Q}[\log(\sigma(f_{\theta_Q}(I,  \nonumber\\
    &Q_i) - f_{\theta_Q}(I,Q_j)))],\\
loss_A(\theta_A) = &-\mathbb{E}_{(I, Q_0, A_i, A_j) \sim \mathcal{D}^\mathrm{HF}_A}[\log(\sigma(g_{\theta_A}(I,  \nonumber\\
    &Q_0, A_i) -g_{\theta_A}(I, Q_0, A_j)))],
\end{align}
where $f$ is the question reward model parameterized by $\theta_Q$ that takes an image $I$ and a question $Q$ as input, and $g$ is the answer reward model parameterized by $\theta_A$, provided with an extra answer $A$. Both $f_{\theta_Q}(I, Q)$ and $g_{\theta_A}(I, Q, A)$ are scalar scores for given instructions. $\sigma$ is the sigmoid function, and $\mathcal{D}^\mathrm{HF}_Q$, $\mathcal{D}^\mathrm{HF}_A$ are the question and answer dataset, respectively.

\subsubsection{Instruction Data Filtration}

Leveraging the reward models well-aligned with human preferences, we implement a two-stage filtration on large-scale datasets, acquiring a substantially smaller yet high-quality dataset. Initially, a visual instruction dataset is synthesized, with each image accompanied by one question and three answers, formally represented as $\mathcal{D} = \{(I^n, Q^n_0, A^n_1, A^n_2, A^n_3)\}_{n \in N}$, where $N=\{1, ..., n_0\}$ is the index set of the full dataset.

In the first filtration stage, we rank all instances in $\mathcal{D}$ based on their question quality using the question reward model $f_{\theta_Q}$, defined as $r^n_Q = f_{\theta_Q}(I^n, Q^n_0)$. Retaining the top $\alpha\%$ of instances yields dataset $\mathcal{D}^{\prime} = \{(I^{n^{\prime}}, Q^{n^{\prime}}_0, A^{n^{\prime}}_1, A^{n^{\prime}}_2, A^{n^{\prime}}_3)\}_{n^{\prime} \in N^{\prime}}$, where $N^{\prime}$ is a subset of $N$, with size $\left\vert N^{\prime} \right\vert = \alpha\% \cdot \left\vert N \right\vert$. The second stage concentrates on evaluating answer quality. For each image in $\mathcal{D}^{\prime}$, the answer reward model $g_{\theta_A}$ assigns scores to the three answer options, computed as $r^{n^{\prime}}_{A, i} = g_{\theta_A}(I^{n^{\prime}}, Q^{n^{\prime}}_0, A^{n^{\prime}}_i)$ for $i \in \{1, 2, 3\}$. The answer with the highest score, denoted as $A^{n^{\prime}}_{\hat{i}}$, is selected for each image. Subsequently, these selected image-question-answer triplets undergo a ranking process analogous to the first stage, where the top $\beta\%$ instances based on answer scores are preserved, leading to the final dataset $\mathcal{D}^{\prime\prime} = \{(I^{n^{\prime\prime}}, Q^{n^{\prime\prime}}_0, A^{n^{\prime\prime}}_{\hat{i}})\}_{n^{\prime\prime} \in N^{\prime\prime}}$, where $N^{\prime\prime} \subset N^{\prime}$ and $\left\vert N^{\prime\prime} \right\vert = \beta\% \cdot \left\vert N^{\prime} \right\vert$. A notable exception is the \emph{detailed description} instructions, which bypass the initial question-based filtration. To maintain consistency in the overall data scale, a direct sampling rate of $\alpha\% \cdot \beta\%$ is applied to this category in the second stage. This two-stage process achieves a sampling rate of $\alpha\% \cdot \beta\%$ relative to the full dataset, effectively retaining instructions best aligned with human preferences in both question and answer domains.

\subsection{LLM Characteristic Alignment} \label{method:LLM_align}

\subsubsection{Inner LLM Rewrite} To harmonize instructions with the writing style of the inner LLM, we finally introduce an LLM alignment step. This alignment comprises two sequential phases: rewriting and review. During the rewriting phase, the inner LLM is tasked with modifying both question and answer components to conform to its characteristic writing style while maintaining semantic equivalence. Accompanying each revised instruction is an explanatory note detailing the modifications, thereby enhancing the process's robustness and interpretability. Crucially, the LLM is instructed to maintain the original phrasing if it already adheres to the LLM's stylistic expectations, mitigating the risk of compromising high-quality instructions.

\subsubsection{Inner LLM Review} In the subsequent review phase, both the original and revised instructions are presented to the LLM. The model is prompted to determine whether the revised content more closely aligns with its writing style while satisfying the semantics preserving demand. This serves as an extra stage of quality assurance. To encourage logical decision-making, the LLM is also required to provide a rationale for its selection. By refining instruction formats to better suit the target MLLM, this alignment process ultimately enhances training efficiency.

\section{Experiments}

\subsection{Setting and Benchmark} \label{exp:setting}

\begin{table*}
    \centering
    \resizebox{0.85\textwidth}{!}{%
    \renewcommand{\arraystretch}{1.25}
    \begin{tabular}{lcc||cccc||cccc}
    \Xhline{1.5pt}
    \multirow{2}{*}{\textbf{\Large Method}} & \multirow{2}{*}{\textbf{\Large LLM}}   & \multirow{2}{*}{\textbf{\Large SMI}} & \multicolumn{4}{c||}{\texttt{VQA Datasets}} & \multicolumn{4}{c}{\texttt{Instruction-Following Benchmarks}} \\
    \Xcline{4-7}{0.2pt} \Xcline{8-11}{0.2pt}
    & & &  \textbf{VQA\textsuperscript{v2}}  & \textbf{VizWiz} & \textbf{SQA\textsuperscript{I}} & \textbf{VQA\textsuperscript{T}}  & \textbf{MME} & \textbf{MMB}  & \textbf{LLaVA\textsuperscript{W}} & \textbf{MM-Vet} \\
    
    \Xhline{1.5pt}
    InstructBLIP~\cite{ref:InstructBLIP} & Vicuna-7B  & 158K & 76.5 & 34.5  & 60.5  & 50.1   & 1251.8 & 36.0  & 56.2  & 26.2    \\
    IDEFICS-9B~\cite{ref:idefics} & LLama-7B & - & 50.9 & 35.5  & 56.7  & 25.9   & 764.0 & 48.2  & \textbf{58.2}  & 27.4    \\
    
    Qwen-VL~\cite{ref:Qwen-VL} & Qwen-7B  & - & \textbf{78.8} & 35.2  & \textbf{67.1}  & \textbf{63.8}   & 1185.4 & 38.2  & 37.8  & \textbf{31.6}    \\
    
    \Xhline{1pt}
    LLaVA1.5$^{\dag}$~\cite{ref:llava1_5} & Vicuna-7B  & 158K & \cellcolor{myorange}75.8 & \cellcolor{myred}\textbf{47.1}  & \cellcolor{myorange}66.1  & \cellcolor{myorange}54.6   & \cellcolor{myorange}1396.9 & \cellcolor{myred}\textbf{62.2}  & \cellcolor{myorange}43.2  & \cellcolor{myorange}27.3 \\
    \rowcolor{gray!40}
    \textbf{\method{}$^{\dag}$ (Ours)} & Vicuna-7B  & 14K & \cellcolor{myred}75.9 & \cellcolor{myorange}45.6 & \cellcolor{myred}\textbf{66.8} & \cellcolor{myred}55.1 & \cellcolor{myred}\textbf{1448.0} & \cellcolor{myorange}61.1 & \cellcolor{myred}50.2 & \cellcolor{myred}28.1  \\
    \rowcolor{lightblue} \textbf{Improvement. $ \uparrow$}  & \textbf{-} & \textbf{-} &  \textbf{+0.1} & \textbf{-}  & \textbf{+0.7}  & \textbf{+0.5}   & \textbf{+51.1} & \textbf{-}  & \textbf{+7.0}  & \textbf{+0.8}  \\
    \Xhline{1pt}\\

    \Xhline{1.5pt}
    
    InstructBLIP~\cite{ref:InstructBLIP} & Vicuna-13B & 158K & 74.8 & 33.4  & 63.1  & 50.7   & 1212.8 & 34.2  & 50.1  & 25.6    \\
    Shikra~\cite{ref:Shirka} & Vicuna-13B & 156K & \textbf{77.4} & -  & -  & -   & - & 58.8  & -  & -  \\
    
    \Xhline{1pt}
    LLaVA1.5$^{\dag}$ (158K) ~\cite{ref:llava1_5} & Vicuna-13B  & 158K & \cellcolor{myorange}\textbf{77.4} & \cellcolor{myred}\textbf{53.9} & \cellcolor{myorange}69.5 & \cellcolor{myorange}\textbf{58.8}  & \cellcolor{myorange}1471.1 & \cellcolor{myorange}63.8  & \cellcolor{myorange}43.6  & \cellcolor{myorange}29.6  \\
    \rowcolor{gray!40}
    \textbf{\method{}$^{\dag}$ (14K) (Ours)} & Vicuna-13B  & 14K & \cellcolor{myorange}\textbf{77.4} & \cellcolor{myorange}51.5 & \cellcolor{myred}\textbf{70.0} & \cellcolor{myorange}\textbf{58.8} & \cellcolor{myred}\textbf{1495.9} & \cellcolor{myred}\textbf{64.0} & \cellcolor{myred}\textbf{53.0} & \cellcolor{myred}\textbf{31.3}  \\
    \rowcolor{lightblue} \textbf{Improvement. $ \uparrow$}  & \textbf{-} & \textbf{-} &  \textbf{-} & \textbf{-}  & \textbf{+0.5}  & \textbf{-}   & \textbf{+24.8} & \textbf{+0.2}  & \textbf{+9.4}  & \textbf{+1.7}  \\
    \Xhline{1pt}
    \end{tabular}
    }    
    \caption{Comparison with Baselines on 8 benchmarks. For fair comparison, we refer $^{\dag}$ to the model trained on our automatically synthesized instructions. SMI indicates the number of synthetic multimodal instruction samples. }
    \label{tab:main_results}
\end{table*}

\subsubsection{Experimental Setting}
To obtain the synthetic visual instruction set, we first generated instructions for 158K images from the MSCOCO~\cite{ref:mscoco} dataset, using CogVLM-17B~\cite{wang2023cogvlm} with 4-bit quantization. 
Our curation pipeline was applied to this synthetic data, with 30\% sampling rate for both questions and answers, resulting in a reduced dataset, \dataset{}, constituting 9\% of the original samples. Combined with the image-caption and short VQA data from the LLaVA-1.5-665K~\cite{ref:llava1_5} dataset, \dataset{} was subsequently used to fine-tune the LLaVA-1.5~\cite{ref:llava1_5} model, initialized with weights from the previous stage pre-trained on caption datasets. We compared the resulting model against the baseline of LLaVA-1.5 fine-tuned on the entire synthetic instruction dataset with caption and VQA data, as well as other state-of-the-art methods.

\subsubsection{Benchmarks}
This evaluation suite comprises academic VQA benchmarks, including VQAv2~\cite{ref:VQA}, VizWiz~\cite{ref:VizWiz}, ScienceQA (IMG)~\cite{ref:SQA}, and TextVQA~\cite{ref:VQAT}. Additionally, to measure instruction-following abilities, we employ MME-Perception~\cite{ref:MME}, MMBench~\cite{ref:MMB}, LLaVA-Bench-in-the-Wild~\cite{ref:LLaVA}, and MM-Vet~\cite{ref:MM-Vet}. These benchmarks collectively provide a robust assessment of multimodal understanding, open-ended conversation, and other relevant skills.

\subsection{Overall Performance}
Table \ref{tab:main_results} presents the main experiment results. Our method outperforms the baselines on the majority of benchmarks while achieving highly competitive results on the remaining ones. A key finding is the significant reduction in synthetic instruction data required for our model to achieve such strong performance. Specifically, our model was fine-tuned on only 9\% of the synthetic instruction data compared to the full-data baseline using the same model architecture.
This outcome strongly suggests that our proposed alignment pipeline not only effectively compresses data size, but also enhances data quality and training efficiency.

\subsection{Ablation Studies} \label{section4:ablation}

\subsubsection{Ablation on Human Alignment}

To investigate the impact of human knowledge alignment, we conducted an ablation study comparing our human-preference-based data filtration method to a random sampling baseline. The results, presented in Table \ref{tab:ablation_human}, demonstrate the consistent superiority of our proposed method. Notably, the performance gains are particularly pronounced on instruction-following benchmarks. For instance, on MME~\cite{ref:MME}, we observe a significant improvement from 1405.3 to 1448.0, while on LLaVA-Bench-In-the-Wild~\cite{ref:LLaVA}, the score rises from 41.6 to 50.2. These substantial gains strongly suggest that aligning data with human preferences is a critical factor in enhancing the model's capacity to generate responses that closely correspond to queries and visual context. This alignment ultimately contributes to the production of human-aligned outputs that effectively fulfill user intents.

\subsubsection{Ablation on LLM Alignment}

We additionally performed an ablation study to investigate the individual contributions of the rewriting and review stages in our LLM alignment process. Specifically, we sequentially removed each stage to assess its impact on model performance. As detailed in Table \ref{tab:ablation_LLM}, the combined application of LLM rewriting and review consistently outperforms the baseline method with solely human-aligned filtration. However, omitting the review stage and directly adopting refined instructions yields unstable results. While achieving a peak accuracy of 67.4 on ScienceQA (IMG)~\cite{ref:SQA}, a substantial performance degradation from 1431.3 to 1408.8 is also observed on the MME~\cite{ref:MME} benchmark. These results indicate that LLM-rewritten instructions are not reliably improved and can potentially compromise the quality of originally superior data. Consequently, the subsequent review stage is essential for identifying and rectifying such issues, thereby enhancing the overall stability and effectiveness of our LLM alignment approach.

\subsection{In-Depth Analysis}
We also conducted a series of experiments to further analyze the effectiveness of \method{}.

\begin{table}[t]
\centering
\resizebox{0.7\columnwidth}{!}{
\renewcommand{\arraystretch}{1.25}
\begin{tabular}{l||cccc}
\Xhline{1.25pt}
\multirow{2}{*}{\textbf{\Large{Data Selection}}}  & \multicolumn{4}{c}{\texttt{Benchmarks}}  \\
\Xcline{2-5}{0.25pt}
& \textbf{VizWiz} & \textbf{SQA\textsuperscript{I}} & \textbf{MME} & \textbf{LLaVA\textsuperscript{W}}  \\
\Xhline{1.25pt}
Random (14K)    & 44.5 & 66.5 & 1405.3 & 41.6  \\
\rowcolor{gray!40}
Ours (14K)      & \textbf{45.6} & \textbf{66.8} & \textbf{1448.0} & \textbf{50.2}  \\
\Xhline{1.25pt}
\end{tabular}
}
\caption{Comparison between our human-preference-aligned filtration and random compression strategy.}
\label{tab:ablation_human}
\end{table}
\begin{table}[t]
\centering
\resizebox{0.9\columnwidth}{!}{
\renewcommand{\arraystretch}{1.25}
\begin{tabular}{l||ccc||cccc}
\Xhline{1.25pt}
\multirow{2}{*}{\textbf{\Large{Method}}}  & 
\multicolumn{3}{c||}{\texttt{Stages}} & \multicolumn{4}{c}{\texttt{Benchmarks}}  \\
\Xcline{2-4}{0.25pt} \Xcline{5-8}{0.25pt}
& \textbf{Human} & \textbf{Rewrite} & \textbf{Review} & \textbf{SQA\textsuperscript{I}} & \textbf{VQA\textsuperscript{T}} & \textbf{MME} & \textbf{LLaVA\textsuperscript{W}}  \\
\Xhline{1.25pt}
Non-LLM-Aligned & \faCheckCircle  &                &                & 66.3 & 54.8 & 1431.3 & 42.9  \\
LLM Rewrite     & \faCheckCircle  & \faCheckCircle &                & \textbf{67.4} & 54.9 & 1408.8 & 49.0  \\
\rowcolor{gray!40}
Ours            & \faCheckCircle  & \faCheckCircle & \faCheckCircle & 66.8 & \textbf{55.1} & \textbf{1448.0} & \textbf{50.2}  \\
\Xhline{1.25pt}
\end{tabular}%
}
\caption{Abalation results of involving different stages in LLM alignment.}
\label{tab:ablation_LLM}
\end{table}

\subsubsection{Object Hallucination Evaluation}

\begin{table}[t]
\centering
\resizebox{1\columnwidth}{!}{
\renewcommand{\arraystretch}{1.25}
\begin{tabular}{l||ccc||ccc||ccc}
\Xhline{1.25pt}
\multirow{2}{*}{\textbf{\Large{Method}}}  & 
\multicolumn{3}{c||}{\texttt{Adversarial}} & \multicolumn{3}{c||}{\texttt{Popular}} & \multicolumn{3}{c}{\texttt{Random}} \\
\Xcline{2-4}{0.25pt} \Xcline{5-7}{0.25pt} \Xcline{8-10}{0.25pt}
& \textbf{Acc} & \textbf{F1} & \textbf{Yes} & \textbf{Acc} & \textbf{F1} & \textbf{Yes} & \textbf{Acc} & \textbf{F1} & \textbf{Yes}  \\
\Xhline{1.25pt}
LLaVA1.5$^{\dag}$~\cite{ref:llava1_5} & 84.1  & 82.7 & 42.2 & 86.1 & 84.6 & 42.2 & 86.7 & 85.5 & 40.4   \\
\rowcolor{gray!40}
\method{}$^{\dag}$ (Ours)                & \textbf{84.5}  & \textbf{83.4} & 43.4 & \textbf{86.6} & \textbf{85.3} & 41.3 & \textbf{87.1} & \textbf{86.1} & 41.6   \\
\Xhline{1.25pt}
\end{tabular}
}
\caption{Zero-shot object hallucination evaluation results on POPE benchmark.}
\label{tab:pope}
\end{table}

We adopted the POPE \cite{ref:POPE} benchmark to evaluate the models' propensity for hallucination, which utilizes a polling-based query method involving yes/no questions. We report accuracy and F1-scores as primary performance metrics. The proportion of positive responses, denoted as "Yes" ratio, is also included as a reference for analyzing model behaviors. The results, presented in Table \ref{tab:pope}, indicate a consistent superiority of our method over the baseline across all three experimental settings. Notably, despite being trained on only 9\% of the original synthetic instructions, our model demonstrates a slight reduction in hallucination. The results underscores the effectiveness of our pipeline in identifying and eliminating factual inaccuracies through data alignment, which is a critical aspect of reliable LLM performance.

\subsubsection{Human Evaluation}

\begin{figure}[t]
    \centering
    \includegraphics[width=0.9\linewidth]{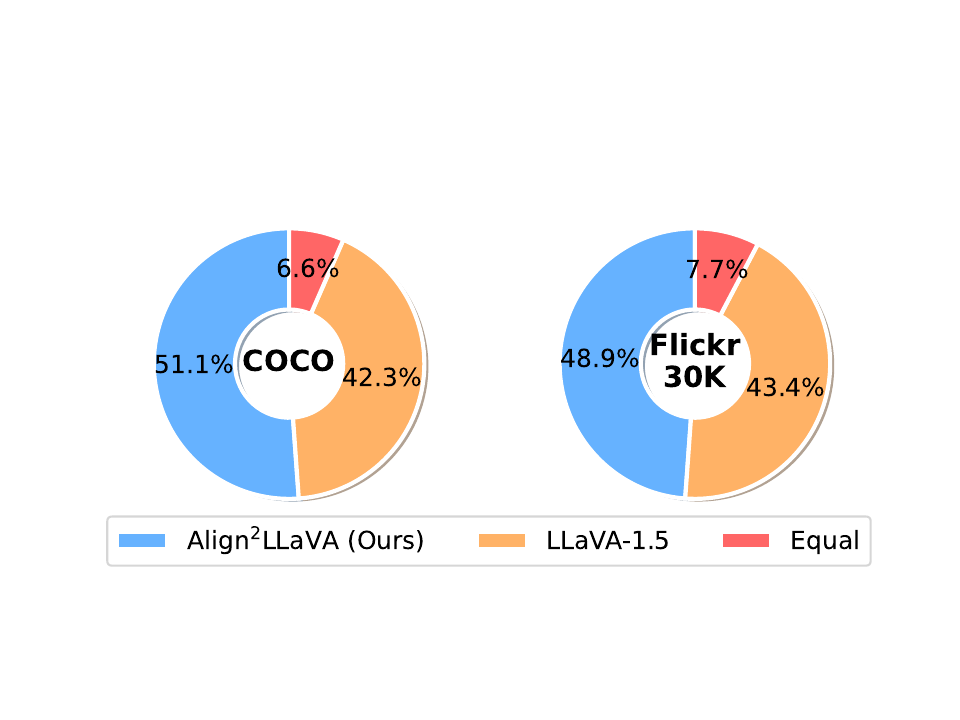}
    \caption{Human evaluation of LLaVA and Align$^2$LLaVA.}
    \label{fig:human_eval}
\end{figure}

To further assess model performance, a human evaluation was conducted on two sets of 50 images randomly sampled from each of the MSCOCO \cite{ref:mscoco} and Flickr30K \cite{ref:flickr30k} datasets. For each image, GPT-4 was prompted to generate questions, which were subsequently refined manually. \method{} model and the baseline LLaVA-1.5 generated responses to these questions, and 7 human evaluators independently compare the quality of the generated responses to each question.
As illustrated in Figure \ref{fig:human_eval}, our method consistently surpasses the baseline model in terms of human preference. These results provide empirical evidence that our approach effectively aligns the model with human expectations, enhancing the model's capacity of producing human-satisfying outputs.

\subsubsection{Transfer to Different LLMs} \label{section4:llm-transfer} 
\begin{figure}[!]
    \centering
    \includegraphics[width=0.7\linewidth]{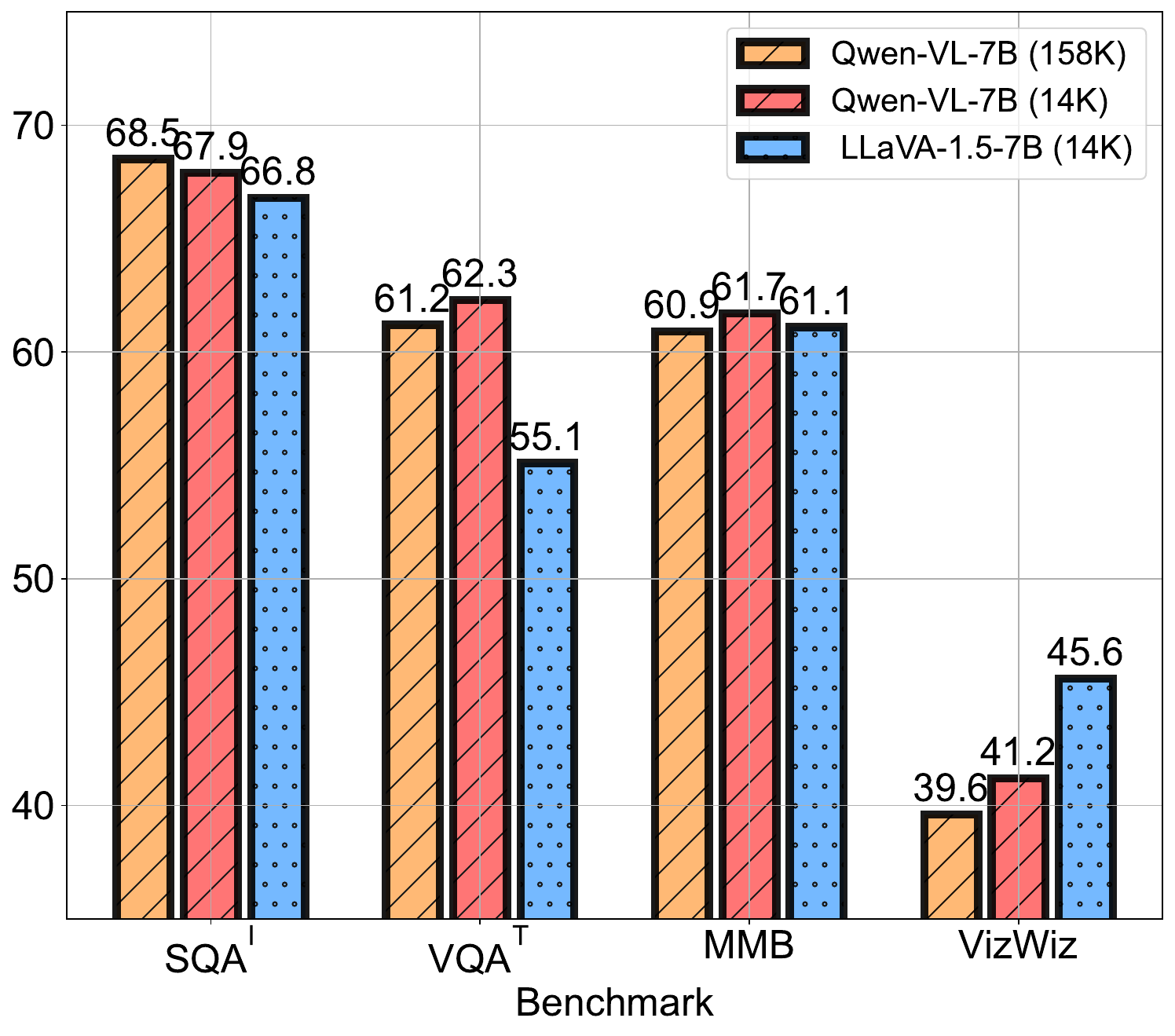}
    \caption{Performance transferring to different LLMs.}
    \label{fig:diff_LLMs}
\end{figure}

We have also explored the transferability of our curated instructions to different LLMs. Beyond the LLaVA-1.5 based on Vicuna-7B, we fine-tuned Qwen-VL-7B~\cite{ref:Qwen-VL}  under identical settings of Section \ref{exp:setting}. As detailed in Figure \ref{fig:diff_LLMs}, on Qwen-7B models, our method achieves comparable performance compared to the counterpart tuned on full 158K synthetic instructions, similar to the findings of fine-tuning Vicuna-7B models. This outcome suggests the generalizability of our \method{} in across various LLM architectures.

\section{Conclusion}

This paper contributes to the advancement of high-quality visual instruction synthesis by introducing a novel filtration pipeline that significantly reduces dataset size. Our pipeline sequentially refines instruction quality through alignment with human preferences and LLM writing style. By applying this pipeline to synthetic data, we demonstrate that the resulting substantially smaller dataset yields performance on par with the full instruction dataset when fine-tuning MLLMs. Our results highlight the potential for improving instruction quality via human-in-the-loop refinement and offer a path towards more efficient MLLM training.
\section*{Limitations}

Our approach, while effective, is subject to certain limitations. Firstly, despite the implementation of unified criteria and rigorous reviews to maintain annotation consistency, inherent subjectivity in human judgment can still introduce discrepancies in human preference data, particularly when multiple annotators are involved. Secondly, while the required amount of human preference data is relatively modest, human annotation remains an essential step in our pipeline. While scaling the synthetic dataset within the current domain does not incur additional annotation costs, extending this method to new domains necessitates the annotation of human preference data in the target domain to enhance the accuracy of the reward models. This potential increase in labor costs associated with adapting our method to novel scenarios should be carefully considered. Furthermore, the acquisition of human preference data in real-world applications may raise privacy concerns. It is imperative to identify and address these privacy issues to ensure the ethical and responsible application of our method.

\bibliography{reference}

\appendix
\section{Training Details} \label{apd:training_detail}

For the balance between speed and precision, all models are trained with BF16 and TF32 enabled, using LoRA with $r=128$ and $\alpha=256$. Model parameters are updated using AdamW optimizer with $\beta_1=0.9$, $\beta_2=0.999$, $\epsilon=1e-8$ and weight decay $0$. For all models, learning rate is set to $2e-5$, with warmup ratio $0.03$ and cosine decay schedule. Reward models are trained on 4 $\times$ RTX 3090s for 5 epochs, using global batch size 64, spending 8 hours for the question model and 9 hours for the answer model. LLaVA-1.5 instruction fine-tuning is completed on 8 $\times$ A100s for 1 epoch with global batch size 128. The training takes 5.5/9.5 hours (7B/13B models) on our aligned dataset and 7/12 hours (7B/13B models) on the full dataset.

\section{Detailed Guidelines for Human Annotation} \label{apd:guideline}

\textbf{Question Annotation Guidelines:}

\begin{itemize}
    \item {\bf Correctness.} Questions are evaluated for their alignment with both image content and general world knowledge. Questions containing inaccuracies, contradictions, or illogical statements are penalized.
    \item {\bf Fluency.} Questions are assessed for grammatical correctness, clarity, and coherence. Ambiguous or poorly structured questions are unacceptable.
    \item {\bf Relevance.} Questions are expected to be highly dependent on the visual information provided in the image. Questions easily answered without image context or those that could be directly inferred from general knowledge receive lower scores. Moreover, questions that inadvertently revealed image details were penalized. To ensure question solvability, answers are required to be definitively obtainable from the image, in conjunction with real-world knowledge.
    \item {\bf Question Distribution (for multi-turn conversation).} In multi-turn dialogues, questions are additionally evaluated based on their diversity in difficulty and subject matter. Ideally, questions should progress from simpler to more complex, covering a broad range of image elements. Repetitive or redundant questions are considered suboptimal.
\end{itemize}

\textbf{Answer Annotation Guidelines:}

\begin{itemize}
    \item {\bf Accuracy.} Answers are evaluated for their correctness and coherence, ensuring alignment with both image content and general knowledge. Responses containing factual errors or contradictions are penalized.
    \item {\bf Completeness.} Answers are expected to provide exhaustive responses to the posed questions, incorporating as much relevant image-derived detail as possible.
    \item {\bf Reasoning.} For questions requiring logical deduction, answers are assessed based on the clarity and correctness of the reasoning process presented. Step-by-step explanations are preferred.
    \item {\bf Relevance.} Answers are expected to be focused on the question at hand, avoiding extraneous information unrelated to the image or query.
\end{itemize}

\section{Adaptation to Multi-Turn Conversations} \label{apd:multiturn}

To accommodate the \emph{conversation} type instructions with multi-turn question-answer pairs, adaptations are necessary. Within each example $(I, Q_0, A_0)$ from the seed dataset, the question and answer components can be decomposed into sequences: $Q_0 = \{q_{0,1}, ..., q_{0,t}\}$ and $A_0 = \{a_{0,1}, ..., a_{0,t}\}$, where $(q_{0,i}, a_{0,i}), i \in \{1, ..., t\}$ represents a single-turn question-answer pair, and the number of turns is determined by the question sequence.

Our approach involves a hybrid strategy: evaluating multi-turn questions holistically, while assessing answers on a per-turn basis. This distinction is grounded in the disparate criteria for these two components. Given the inherent interconnectedness and potential redundancies within multi-turn question sequences, a comprehensive evaluation is essential to capture their overall quality. Conversely, the standards for answer quality are predominantly determined on the current question and image, rendering contextual factors less critical. 
Consequently, isolating answer evaluation to the current turn optimizes computational efficiency and prevents the introduction of irrelevant information.

Specifically, for the question component, human evaluators assess question groups holistically within the same image, and the question reward model also processes entire question groups as input, with individual questions concatenated. For multi-turn answers, each $t$-turn example $(I, Q_0, A_0, A_1, ..., A_l)$ is decomposed into $t$ single-turn examples $(I, q_{0,1}, a_{0,1}, ..., a_{l,1}), ..., (I, q_{0,t}, a_{0,t}, ..., a_{l,t})$ for subsequent processing, which mirrors the single-turn pipeline. The sole deviation occurs in the second filtration stage of human alignment, where the optimal answer is independently selected for each turn, and their average score serve as the overall answer quality metric for the multi-turn instance.

\section{Reward Models Evaluation} \label{apd:reward}

\begin{table}[t]
\centering
\resizebox{0.8\columnwidth}{!}{
\renewcommand{\arraystretch}{1.25}
\begin{tabular}{l||cc||cc}
\Xhline{1.25pt}
\multirow{2}{*}{\textbf{\Large{Method}}}  & 
\multicolumn{2}{c||}{\texttt{Direct Prediction}} & \multicolumn{2}{c}{\texttt{Benchmarks}} \\
\Xcline{2-3}{0.25pt} \Xcline{4-5}{0.25pt}
& \textbf{Question} & \textbf{Answer} & \textbf{MME} & \textbf{LLaVA\textsuperscript{W}} \\
\Xhline{1.25pt}
CLIP Score & 44.6          & 51.5          & 1389.4          & 40.4          \\
\rowcolor{gray!40}
Ours       & \textbf{60.6} & \textbf{65.1} & \textbf{1448.0} & \textbf{50.2} \\
\Xhline{1.25pt}
\end{tabular}
}
\caption{Comparison between CLIP and our reward model.}
\label{tab:reward_model}
\end{table}

To construct question and answer reward models, we adapted the LLaVA-1.5-7B architecture by substituting its language modeling head with a sequence classification head, whose output dimension was set as 1. These models were trained on human preference datasets derived from the first step of our pipeline, comprising 800 question instances and 960 answer instances, and further partitioned into 700/100 and 800/160 training/testing sets. Model training was conducted on the training split, followed by evaluation of prediction accuracy on the test split. 
For comparison, we additionally incorporated a baseline method of CLIP~\cite{ref:CLIP} Score between the image and instruction, which is typical of non-human-feedback evaluation. We also integrated both methods into the data curation pipeline, and subsequently compared the trained LLaVA-1.5 performance.
As detailed in Table \ref{tab:reward_model}, our proposed reward models consistently outperform the CLIP Score baseline in both direct assessment and trained LLaVA-1.5 performance. 
These results underscore the effectiveness of our reward models in capturing human-centric evaluation criteria for both questions and answers.

We have further analyzed some failure cases of our reward model in Table \ref{tab:failure_question} and Table \ref{tab:failure_answer}. It can be found that in reward models' misprediction cases, the two compared samples are often similar in quality, and the model gives close scores to both samples. Although recorded as a prediction error, the choice of either sample have no significant impact on data quality.

\section{Prompts} \label{apd:prompts}

For question generation using text-only GPT4, we directly use the same prompts as LLaVA, and discard the generated answer parts. For answer generation, we adjust the prompts and provide the questions to ensure the model only produces answers. Moreover, since visual-capable GPT-4V and CogVLM directly see the image, we further modify the prompts to accommodate changes in input format. Table \ref{tab:prompt_question_complex_reasoning}, Table \ref{tab:prompt_question_conversation}, Table \ref{tab:prompt_answer_complex_reasoning}, Table \ref{tab:prompt_answer_multi}, Table \ref{tab:prompt_answer_detail} show detailed prompts used for generation. Additionally, in the LLM alignment step, our prompts for rewriting and review are detailed in Table \ref{tab:prompt_llm}.

\section{Additional Examples} \label{apd:examples}

We exhibit more examples of different categories to demonstrate our alignment pipeline, as shown in Table \ref{tab:example_reasoning}, Table \ref{tab:example_detail} and Table \ref{tab:example_conv}.

\begin{table*}[!]
    \centering
    \begin{tabular}{p{0.5\linewidth}|p{0.5\linewidth}}
        \Xhline{1.25pt}
        \multicolumn{2}{l}{\texttt{Question Failure Case 1}} \\
        \Xhline{0.25pt}
        \multicolumn{2}{c}{\includegraphics[width=0.4\linewidth]{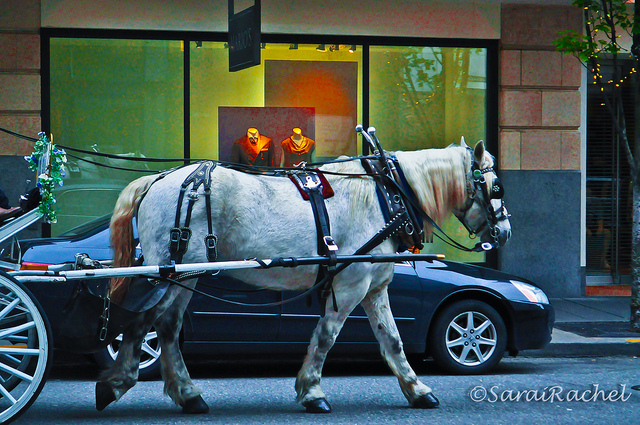}} \\
        \Xhline{0.25pt}
        \textbf{Question 1 (Human Preferred):}
        
        What tradition or practice could be observed in this city street?
        &\textbf{Question 2:}

        Given the presence of both a horse and a car, what kind of transportation conflict might exist? \\
       \textbf{ RM Score:} 4.3
        & \textbf{RM Score:} 4.8 \\
        \Xhline{1.25pt}
        \multicolumn{2}{c}{}\\

        \Xhline{1.25pt}
        \multicolumn{2}{l}{\texttt{Question Failure Case 2}} \\
        \Xhline{0.25pt}
        \multicolumn{2}{c}{\includegraphics[width=0.4\linewidth]{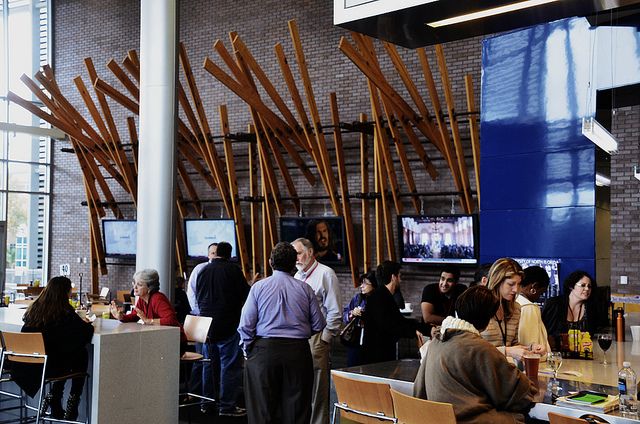}} \\
        \Xhline{0.25pt}
        \textbf{Question Group 1 (Human Preferred):}
        
        1. How many people are visible in this restaurant scene?
        
        2. Where are the people located? Are they standing or sitting?
        
        3. What types of objects can you identify in this image besides people?
        
        4. What kind of venue does it seem to be, based on the setting and objects present?
        
        5. Based on the image, describe the ambiance or atmosphere of the venue.
        &\textbf{Question Group 2:}

        1. How many people can you see in the restaurant image?
        
        2. What are the people doing in the restaurant?
        
        3.What types of objects are visible in the restaurant?
        
        4. How many televisions are there on the wall?
        
        5. What kind of atmosphere does the image suggest about the restaurant? \\
       \textbf{ RM Score:} 7.5
        & \textbf{RM Score:} 8.6 \\
        \Xhline{1.25pt}
        
    \end{tabular}
    \caption{Question reward model failure cases.}
    \label{tab:failure_question}
\end{table*}

\begin{table*}[!]
    \centering
    \begin{tabular}{p{0.5\linewidth}|p{0.5\linewidth}}
        \Xhline{1.25pt}
        \multicolumn{2}{l}{\texttt{Answer Failure Case 1}} \\
        \Xhline{0.25pt}
        \multicolumn{2}{c}{\includegraphics[width=0.4\linewidth]{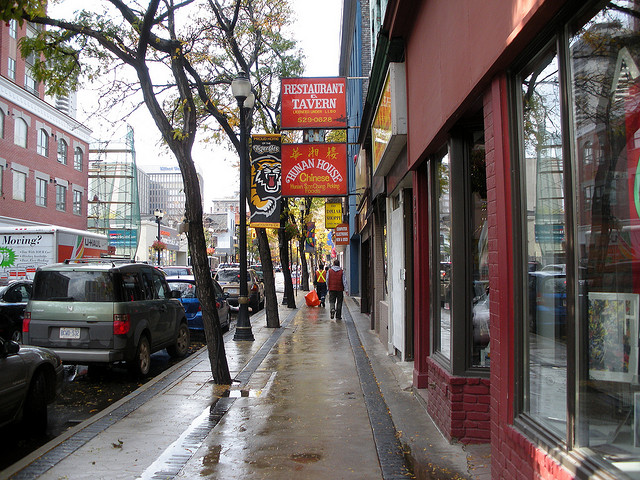}} \\
        \Xhline{0.25pt}
        \multicolumn{2}{l}{\textbf{Question:} Can you see people walking on the city street after the rain in the image?} \\
        \Xhline{0.25pt}
        \textbf{Answer 1 (Human Preferred):}
        
        Yes, there are people walking down the city street after it has been raining. They are likely residents or visitors navigating their way through the city, possibly visiting the restaurants and shops along the street.
        &\textbf{Answer 2:}

        Yes, people can be seen walking down the city street following the rain, likely resuming their activities now that the weather has improved. \\
       \textbf{ RM Score:} 13.3
        & \textbf{RM Score:} 13.6 \\
        \Xhline{1.25pt}
        \multicolumn{2}{c}{}\\

        \Xhline{1.25pt}
        \multicolumn{2}{l}{\texttt{Answer Failure Case 2}} \\
        \Xhline{0.25pt}
        \multicolumn{2}{c}{\includegraphics[width=0.4\linewidth]{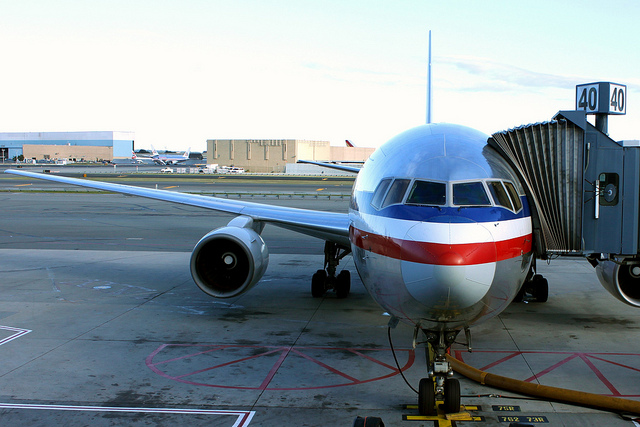}} \\
        \Xhline{0.25pt}
        \multicolumn{2}{l}{\textbf{Question:} What can be inferred about the airplane's current activity?} \\
        \Xhline{0.25pt}
        \textbf{Answer 1 (Human Preferred):}
        
        Based on the details available in the image, it can be inferred that the airplane is likely either loading or unloading passengers at the airport terminal. The presence of a passenger ramp attached to the airplane typically indicates this kind of activity. The process might involve passengers boarding the plane for an upcoming flight or disembarking after a concluded journey. Additionally, while this process takes place, the airplane's cargo could also be being loaded or unloaded, and it might be undergoing routine checks and refueling in preparation for its next journey.
        &\textbf{Answer 2:}

        It can be inferred that the airplane is currently engaged in the process of boarding or deboarding passengers. The connected jetway suggests that there is active interaction between the airplane and the airport terminal, indicating the beginning or conclusion of a flight. The positioning at the gate also implies preparations or finalizations are being made for a flight, which could include loading or unloading of luggage, refueling, and necessary checks by the ground crew before departure or after arrival. \\
       \textbf{ RM Score:} 12.1
        & \textbf{RM Score:} 12.7 \\
        \Xhline{1.25pt}
        
    \end{tabular}
    \caption{Answer reward model failure cases.}
    \label{tab:failure_answer}
\end{table*}

\begin{table*}[!]
    \centering
    \begin{tabular}{p{\linewidth}}
        \Xhline{1.25pt}
        \texttt{Complex Reasoning} \\
        \Xhline{0.25pt}
        \textbf{GPT-4}
        
        You are an AI visual assistant that can analyze a single image. You receive five sentences, each describing the same image you are observing. In addition, specific object locations within the image are given, along with detailed coordinates. These coordinates are in the form of bounding boxes, represented as (x1, y1, x2, y2) with floating numbers ranging from 0 to 1. These values correspond to the top left x, top left y, bottom right x, and bottom right y.

        The task is to use the provided caption and bounding box information, create a plausible question about the image, and provide the answer in detail.
        
        Create complex questions beyond describing the scene.
        To answer such questions, one should require first understanding the visual content, then based on the background knowledge or reasoning, either explain why the things are happening that way, or provide guides and help to user's request.  Make the question challenging by not including the visual content details in the question so that the user needs to reason about that first.
        
        Instead of directly mentioning the bounding box coordinates, utilize this data to explain the scene using natural language. Include details like object counts, position of the objects, relative position between the objects.  
        
        When using the information from the caption and coordinates, directly explain the scene, and do not mention that the information source is the caption or the bounding box.  Always answer as if you are directly looking at the image. 
        
        \textless captions\textgreater \textless bounding boxes\textgreater \\
        \textbf{GPT-4V/CogVLM}

        You are an AI visual assistant that can analyze a single image.

        The task is to create a plausible question about the image, and provide the answer in detail.
        
        Create complex questions beyond describing the scene.
        To answer such questions, one should require first understanding the visual content, then based on the background knowledge or reasoning, either explain why the things are happening that way, or provide guides and help to user's request.  Make the question challenging by not including the visual content details in the question so that the user needs to reason about that first.
        
        Include details like object counts, position of the objects, relative position between the objects.  
        
        Do not mention anything from the prompt in your response. 
        
        \textless image\textgreater \\
        \Xhline{1.25pt} \\
    \end{tabular}
    \caption{Prompts for \textit{complex reasoning} question generation.}
    \label{tab:prompt_question_complex_reasoning}
\end{table*}

\begin{table*}[!]
    \centering
    \begin{tabular}{p{\linewidth}}
        \Xhline{1.25pt}
        \texttt{Conversation} \\
        \Xhline{0.25pt}
        \textbf{GPT-4}

        You are an AI visual assistant, and you are seeing a single image. What you see are provided with five sentences, describing the same image you are looking at. Answer all questions as you are seeing the image.

        Design a conversation between you and a person asking about this photo. The answers should be in a tone that a visual AI assistant is seeing the image and answering the question.
        Ask diverse questions and give corresponding answers.
        
        Include questions asking about the visual content of the image, including the object types, counting the objects, object actions, object locations, relative positions between objects, etc. Only include questions that have definite answers:
        (1) one can see the content in the image that the question asks about and can answer confidently;
        (2) one can determine confidently from the image that it is not in the image.
        Do not ask any question that cannot be answered confidently.
        
        Also include complex questions that are relevant to the content in the image, for example, asking about background knowledge of the objects in the image, asking to discuss about events happening in the image, etc. Again, do not ask about uncertain details.
        Provide detailed answers when answering complex questions. For example, give detailed examples or reasoning steps to make the content more convincing and well-organized.  You can include multiple paragraphs if necessary. 
        
        \textless captions\textgreater \textless bounding boxes\textgreater \\
        \textbf{GPT-4V/CogVLM}

        You are an AI visual assistant, and you are seeing a single image. Answer all questions according to the image.

        Design a conversation between you and a person asking about this photo. The answers should be in a tone that a visual AI assistant is seeing the image and answering the question.
        Ask diverse questions and give corresponding answers.
        
        Include questions asking about the visual content of the image, including the object types, counting the objects, object actions, object locations, relative positions between objects, etc. Only include questions that have definite answers:
        (1) one can see the content in the image that the question asks about and can answer confidently;
        (2) one can determine confidently from the image that it is not in the image.
        Do not ask any question that cannot be answered confidently.
        
        Also include complex questions that are relevant to the content in the image, for example, asking about background knowledge of the objects in the image, asking to discuss about events happening in the image, etc. Again, do not ask about uncertain details.
        Provide detailed answers when answering complex questions. For example, give detailed examples or reasoning steps to make the content more convincing and well-organized.  You can include multiple paragraphs if necessary. 
        
        \textless image\textgreater \\
        \Xhline{1.25pt}
    \end{tabular}
    \caption{Prompts for \textit{conversation} question generation.}
    \label{tab:prompt_question_conversation}
\end{table*}

\begin{table*}[!]
    \centering
    \begin{tabular}{p{\linewidth}}
        \Xhline{1.25pt}
        \texttt{Complex Reasoning} \\
        \Xhline{0.25pt}
        \textbf{GPT-4}
        
        You are an AI visual assistant that can analyze a single image. You receive five sentences, each describing the same image you are observing. In addition, specific object locations within the image are given, along with detailed coordinates. These coordinates are in the form of bounding boxes, represented as (x1, y1, x2, y2) with floating numbers ranging from 0 to 1. These values correspond to the top left x, top left y, bottom right x, and bottom right y.

        Then, you are given a plausible question about the image. The task is to use the provided caption and bounding box information to provide the answer in detail.
        
        To answer such questions, you should first understanding the visual content, then based on the background knowledge or reasoning, either explain why the things are happening that way, or provide guides and help to user's request.
        
        Instead of directly mentioning the bounding box coordinates, utilize this data to explain the scene using natural language. Include details like object counts, position of the objects, relative position between the objects.  
        
        When using the information from the caption and coordinates, directly explain the scene, and do not mention that the information source is the caption or the bounding box.  Always answer as if you are directly looking at the image.
        
        \textless captions\textgreater \textless bounding boxes\textgreater 
        
        Question: \textless question\textgreater 

        ===
        
        Answer: \\
        \textbf{GPT-4V/CogVLM}

        You are an AI visual assistant that can analyze a single image.

        The task is to create a plausible question about the image, and provide the answer in detail.
        
        Create complex questions beyond describing the scene.
        To answer such questions, one should require first understanding the visual content, then based on the background knowledge or reasoning, either explain why the things are happening that way, or provide guides and help to user's request.  Make the question challenging by not including the visual content details in the question so that the user needs to reason about that first.
        
        Include details like object counts, position of the objects, relative position between the objects.  
        
        Do not mention anything from the prompt in your response. 
        
        \textless image\textgreater 
        
        Question: \textless question\textgreater 

        ===
        
        Answer: \\
        \Xhline{1.25pt} \\
    \end{tabular}
    \caption{Prompts for \textit{complex reasoning} answer generation.}
    \label{tab:prompt_answer_complex_reasoning}
\end{table*}

\begin{table*}[!]
    \centering
    \begin{tabular}{p{\linewidth}}
        \Xhline{1.25pt}
        \texttt{Conversation} \\
        \Xhline{0.25pt}
        \textbf{GPT-4 System Message}
        
        You are an AI visual assistant, and you are seeing a single image. What you see are provided with five sentences, describing the same image you are looking at. Answer all questions as you are seeing the image.

        The answers should be in a tone that a visual AI assistant is seeing the image and answering the question.
        
        Provide detailed answers when answering complex questions. For example, give detailed examples or reasoning steps to make the content more convincing and well-organized.  You can include multiple paragraphs if necessary.
        
        \textless captions\textgreater \textless bounding boxes\textgreater \\
        \textbf{GPT-4V/CogVLM System Message}

        You are an AI visual assistant, and you are seeing a single image. Answer all questions according to the image.

        High-quality answers should demonstrate a deep understanding of the visual information, drawing connections between different elements in the image and providing thoughtful insights. Your responses should be clear, concise, and logically structured.
        
        For complex questions, give detailed examples or reasoning steps to make the content more convincing and well-organized.  You can include multiple paragraphs if necessary.
        
        Do not mention anything from the prompt in your response.
        
        \textless image\textgreater \\
        \Xhline{0.25pt}
        \textbf{Turn 1}

        \textless system message\textgreater

        Question: \textless question 1\textgreater 

        ===
        
        Answer: \\
        \Xhline{0.25pt}
        \textbf{Turn 2}

        \textless system message\textgreater

        Question: \textless question 1\textgreater 

        ===
        
        Answer: \textless generated answer 1\textgreater

        ===

        Question: \textless question 2\textgreater 

        ===
        
        Answer: \\
        \Xhline{0.25pt}
        ... \\
        
        \Xhline{1.25pt}
    \end{tabular}
    \caption{Prompts for \textit{conversation} answer generation. Answers of different turns are generated autoregressively.}
    \label{tab:prompt_answer_multi}
\end{table*}

\begin{table*}[!]
    \centering
    \begin{tabular}{p{\linewidth}}
        \Xhline{1.25pt}
        \texttt{Detail Description} \\
        \Xhline{0.25pt}
        \textbf{GPT-4}

        You are an AI visual assistant that can analyze a single image. You receive five sentences, each describing the same image you are observing. In addition, specific object locations within the image are given, along with detailed coordinates. These coordinates are in the form of bounding boxes, represented as (x1, y1, x2, y2) with floating numbers ranging from 0 to 1. These values correspond to the top left x, top left y, bottom right x, and bottom right y.

        Using the provided caption and bounding box information, describe the scene in a detailed manner.
        
        Instead of directly mentioning the bounding box coordinates, utilize this data to explain the scene using natural language. Include details like object counts, position of the objects, relative position between the objects.
        
        When using the information from the caption and coordinates, directly explain the scene, and do not mention that the information source is the caption or the bounding box.  Always answer as if you are directly looking at the image.
        
        \textless captions\textgreater \textless bounding boxes\textgreater \\
        \textbf{GPT-4V/CogVLM}

        You are an AI visual assistant that can analyze a single image.

        Given the image, your task is to describe the scene in a detailed manner.
        
        Include details like object counts, position of the objects, relative position between the objects.
        
        Do not mention anything from the prompt in your response.
        
        \textless image\textgreater \\
        \Xhline{1.25pt}
    \end{tabular}
    \caption{Prompts for \textit{detail description} answer generation.}
    \label{tab:prompt_answer_detail}
\end{table*}

\begin{table*}[!]
    \centering
    \begin{tabular}{p{\linewidth}}
        \Xhline{1.25pt}
        \texttt{LLM Alignment} \\
        \Xhline{0.25pt}
        \textbf{Rewrite}
        
        Given the following Question and Answer, you are required to revise the Question and Answer in your writing style without changing the semantic meaning. If you think the  original question and original answer are clear and consistent with your writing style, just leave it unchanged. The response should contain just the revised question and revised answer and the explanation of revision, formatted as: 'Revised Question:', 'Revised Answer:' , and 'Explanation:'.
        
        Question: \textless original question\textgreater
        
        Answer: \textless original answer\textgreater \\
        \Xhline{0.25pt}
        \textbf{Review}
        
        Given the following Original Question, Revised Question,  Original Answer, and the Revised Answer, if the Revised Question or Revised Answer is inconsistent with your writing style, or deviates from the semantics of Revised Question or Original Answer, or adds or lacks any information, output "There is something wrong with the Revised Question or Revised Answer." Otherwise, output "The Revised Question and Revised Answer are fine." After making your decision, please provide a detailed explanation of your reasoning.
        
        Original Question: \textless original question\textgreater
        
        Original Answer: \textless original answer\textgreater
        
        Revised Question: \textless revised question\textgreater
        
        Revised Answer: \textless revised answer\textgreater  \\
        
        \Xhline{1.25pt}
    \end{tabular}
    \caption{Prompts for LLM alignment.}
    \label{tab:prompt_llm}
\end{table*}

\begin{table*}[!]
    \centering
    \begin{tabular}{p{\linewidth}}
        \Xhline{1.25pt}
        \texttt{Complex Reasoning} \\
        \Xhline{0.25pt}
        \multicolumn{1}{c}{\includegraphics[width=0.3\linewidth]{pics/apd_example_reasoning.jpg}} \\
        \Xhline{0.25pt}
        \texttt{Human Alignment (Question)} \\
        \Xhline{0.25pt}

        \textbf{Question 1 (Accepted):} What possible purpose could the handheld object in the image serve for the person?

        \textbf{Question 2:} How does the inclusion of traffic light in public exhibits reflect societal values and priorities concerning urban planning and safety? \\
        \Xhline{0.25pt}
        \texttt{Human Alignment (Answer)} \\
        \Xhline{0.25pt}

        \textbf{Question:} What possible purpose could the handheld object in the image serve for the person?
        
        \textbf{Answer 1 (Accepted):} The handheld object is not distinctly clear in the image, but given the context of the person walking past a traffic light in what appears to be \textcolor{blue}{an indoor setting}, it is possible that the object could serve as a guide or an \textcolor{blue}{educational tool}. If we imagine a scenario, it might be an interactive exhibit or part of a demonstration to teach about road safety or traffic signals. The individual might be using it to engage with the exhibit, perhaps by pressing buttons to change the lights or to simulate crossing the street while following traffic signals. Such educational setups are often found in \textcolor{blue}{museums}, especially in those \textcolor{blue}{tailored towards children}, to provide a hands-on learning experience about everyday situations and safety.

        \textbf{Answer 2:} The handheld object the person is carrying appears to be a handbag. Handbags serve multiple purposes, such as carrying personal belongings, essential items, and accessories. For example, the person might have their wallet, keys, mobile phone, and other items in the handbag to keep them organized and easily accessible while walking past the traffic light, be it in \textcolor{red}{a busy urban setting} like Walnut Depot in the Bronx, or elsewhere. Additionally, handbags can also serve as a fashion accessory or a style statement, complementing the person's outfit and overall appearance.
        
        \textbf{Answer 3:} In the image, the person walking past the traffic light is carrying a handbag. This handheld object serves multiple purposes for the individual. It is likely being used to carry personal belongings, such as wallet, keys, mobile phone, or other essentials while they are out walking. It can also serve as a fashion accessory, complimenting the individual's overall outfit. Considering the location, where they are walking past a traffic light in a city setting, and \textcolor{red}{the presence of a bus in the background}, it's plausible that this person is commuting and the handbag might carry items required for their journey or destination, such as books, electronics, or work-related items. \\
        \Xhline{0.25pt}
        \texttt{LLM Rewriting} \\
        \Xhline{0.25pt}
        \textbf{Revised Question:} What could be the potential use of the handheld object the person in the image is interacting with?
        
        \textbf{Revised Answer:} The handheld object appears to be an interactive exhibit or educational tool, possibly found in a museum or a child-oriented educational setting. The person might be using it to learn about road safety and traffic signals by pressing buttons to simulate crossing the street while following traffic signals or change the lights. This type of hands-on learning experience is designed to teach children and adults about everyday situations and safety in a fun and engaging way. \\
        \Xhline{1.25pt}
    \end{tabular}
    \caption{An alignment example of \emph{complex reasoning} type instruction.}
    \label{tab:example_reasoning}
\end{table*}

\begin{table*}[!]
    \centering
    \begin{tabular}{p{\linewidth}}
        \Xhline{1.25pt}
        \texttt{Detail Description} \\
        \Xhline{0.25pt}
        \multicolumn{1}{c}{\includegraphics[width=0.4\linewidth]{pics/apd_example_detail.jpg}} \\
        \Xhline{0.25pt}
        \texttt{Human Alignment} \\
        \Xhline{0.25pt}
        \textbf{Answer 1:} A red city bus is traveling down a busy city street, surrounded by parked cars on both sides. The bus is moving between the vehicles and takes up a significant portion of the available street space. 

        \textcolor{red}{In addition to the parked cars, there are also a few people present in the scene. One person is walking near the middle of the scene, while two other individuals are located towards the right side of the image. The scene captures the typical hustle and bustle of a city street with various modes of transportation and pedestrians going about their daily activities.}

        \textbf{Answer 2 (Accepted):} This photo captures a red single-decker bus in service, designated as route 152, which is heading towards Liverpool Street. The bus has a digital display that shows the route number and destination, and it's driving down a city street lined with Georgian-style erraced houses. The scene is set on a day with clear skies.

        \textcolor{blue}{The road is shared with other vehicles, including parked cars along the curb on the right side.} The architectural uniformity of the terraced houses along the street provides a historical backdrop to the modern activity of urban transportation. The setting suggests a typical urban street scene, showcasing the integration of public transport within a residential area.
        
        \textbf{Answer 3:} This photo depicts a red bus on a street lined with parked cars and buildings. The bus has a destination sign indicating "Liverpool Street" and \textcolor{red}{a route number of "153"}. The location appears to be in an urban area, likely in the UK, as indicated by the style of the bus and the street architecture, which includes terraced houses with brick facades. The street is relatively narrow, and the bus is driving in what seems to be a typical residential or mixed-use area. The sky is partly cloudy, suggesting it might be either late afternoon or early morning. \\
        \Xhline{0.25pt}
        \texttt{LLM Rewriting} \\
        \Xhline{0.25pt}
        \textbf{Revised Answer:} This photo appears to have been captured in a residential area in a city, possibly in the UK. The red single-decker bus with the route number 152 and destination of Liverpool Street suggests that the bus is in service and heading towards the Liverpool Street station. The clear skies and the daytime suggest that the photo was taken during daylight hours. The Georgian-style terraced houses on either side of the road suggest that the area has a historical background. The presence of other vehicles, including parked cars along the curb on the right side, indicates that the road is shared with other transportation modes. Overall, the photo showcases the integration of public transport within a residential area. \\
        \Xhline{1.25pt}
    \end{tabular}
    \caption{An alignment example of \emph{detail description} type instruction.}
    \label{tab:example_detail}
\end{table*}

\begin{table*}[!]
    \centering
    \small
    \begin{tabular}{p{0.5\linewidth}|p{0.5\linewidth}}
        \Xhline{1.25pt}
        \multicolumn{2}{l}{\texttt{Conversation}} \\
        \Xhline{0.25pt}
        \multicolumn{2}{c}{\includegraphics[width=0.4\linewidth]{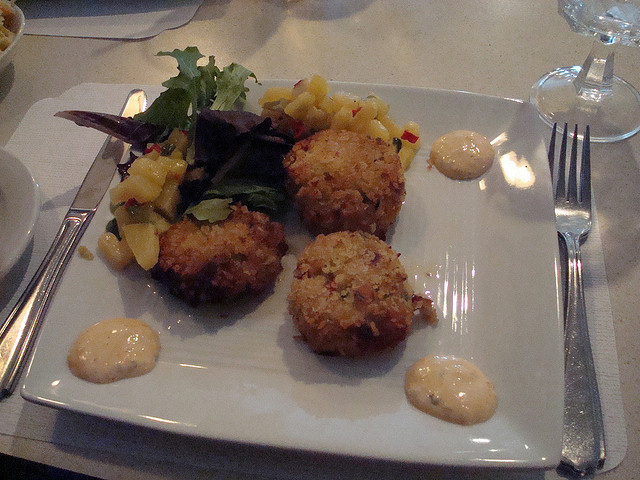}} \\
        \Xhline{0.25pt}
        \multicolumn{2}{p{\linewidth}}{\texttt{Human Alignment (Question)}} \\
        \Xhline{0.25pt}
        \textbf{Questions 1:}
        
        What is the main dish on the plate?
        
        Is the meal on a plate or a tray?
        
        What is the state of the meal?
        
        Can you give an example of how to create a delicious \textcolor{red}{homemade pizza}?
        & \textbf{Questions 2 (Accepted):}
        
        What is shown in the image?

        Where is the dipping sauce located on the plate?

        Are there any utensils visible in the image?

        What dining setting does this image suggest? \\
        \Xhline{0.25pt}
        \multicolumn{2}{p{\linewidth}}{\texttt{Human Alignment (Answer)}} \\
        \Xhline{0.25pt}
        \multicolumn{2}{p{\linewidth}}{
        \textbf{Question:} What is shown in the image? 

        \textbf{Answer 1:} The image shows a plated dish of three crab cakes, accompanied by a small side salad with mixed greens and a mango salsa. There are also \textcolor{red}{two} dollops of a creamy dipping sauce on opposite sides of the plate.

        \textbf{Answer 2 (Accepted):} The image shows a plate of crab cakes, a popular seafood dish. The crab cakes are served with pineapple salsa, creamy sauce, and fresh greens. \newline

        \textbf{Question:} Where is the dipping sauce located on the plate?

        \textbf{Answer 1:} The dipping sauce is located at \textcolor{red}{two spots} on the plate, one on the left and one on the right, symmetrically placed near the edges.

        \textbf{Answer 2 (Accepted):} The dipping sauce is located in \textcolor{blue}{three small dollops} on the plate. They are positioned \textcolor{blue}{around the crab cakes}. \newline

        \textbf{Question:} Are there any utensils visible in the image?

        \textbf{Answer 1 (Accepted):} Yes, there is \textcolor{blue}{a fork visible to the right} of the plate, along with \textcolor{blue}{a knife to the left}. These utensils suggest that the dish is ready to be eaten.

        \textbf{Answer 2:} Yes, there are utensils visible in the image. You can see \textcolor{red}{a fork and a knife on the right side} of the plate. \newline

        \textbf{Question:} What dining setting does this image suggest?

        \textbf{Answer 1:} The image suggests a casual yet refined dining setting. The plating is elegant but not overly formal, and the presence of a wine glass in the background indicates that this could be a meal served in a restaurant or at a home dinner party. The overall atmosphere seems relaxed, making it suitable for a pleasant, leisurely meal.

        \textbf{Answer 2 (Accepted):} The image suggests a dining setting that leans towards a casual to mid-range restaurant. The uncomplicated plate design, paired with the straightforward presentation of the crab cakes, pineapple salsa, and creamy sauce, hints at a relaxed atmosphere. The inclusion of a fork and knife, while standard, further supports the notion of a casual yet attentive dining experience. This type of setting often caters to a broad clientele seeking a satisfying meal without excessive formality.
        } \\
        \Xhline{0.25pt}
        \multicolumn{2}{p{\linewidth}}{\texttt{LLM Rewriting}} \\
        \Xhline{0.25pt}
        \multicolumn{2}{p{\linewidth}}{
        \textbf{Revised Question:} What is depicted in the image?

        \textbf{Revised Answer:} The image depicts a plate of crab cakes, a beloved seafood dish. The crab cakes are accompanied by pineapple salsa, creamy sauce, and a sprig of fresh greens. \newline

        \textbf{Revised Question:} How is the dipping sauce presented on the plate?

        \textbf{Revised Answer:} The dipping sauce is presented in three small dollops on the plate, positioned around the crab cakes. \newline

        \textbf{Revised Question:} What utensils can be seen in the image?

        \textbf{Revised Answer:} A fork is visible to the right of the plate, and a knife is visible to the left. These utensils suggest that the dish is ready to be eaten. \newline

        \textbf{Revised Question:} In what type of dining setting does the image suggest the crab cakes and pineapple salsa are being served?
        
        \textbf{Revised Answer:} The image suggests that the crab cakes and pineapple salsa are being served in a casual to mid-range restaurant. The straightforward presentation of the dish and the inclusion of a fork and knife hint at a relaxed atmosphere. This type of setting caters to a broad clientele seeking a satisfying meal without excessive formality.
        } \\
        
        \Xhline{1.25pt}
    \end{tabular}
    \caption{An alignment example of \emph{conversation} type instruction.}
    \label{tab:example_conv}
\end{table*}

\end{document}